\documentclass[10pt,twocolumn,letterpaper]{article}

\usepackage{cvpr}
\usepackage{times}
\usepackage{epsfig}
\usepackage{graphicx}
\usepackage{amsmath}
\usepackage{amssymb}

\usepackage{indentfirst}
\usepackage{enumitem}
\usepackage[toc,page]{appendix}

\usepackage[pagebackref=true,breaklinks=true,letterpaper=true,colorlinks,bookmarks=false]{hyperref}

\cvprfinalcopy 

\def\cvprPaperID{1289} 

\begin{document}

\title{Bidirectional Attentive Fusion with Context Gating for Dense Video Captioning}

\author{Jingwen Wang$^{\dagger}$\thanks{Work done while Jingwen Wang was a Research Intern with Tencent AI Lab.}\qquad Wenhao Jiang$^{\ddagger}$\renewcommand{\thefootnote}{\fnsymbol{footnote}}\footnotemark[4]\qquad Lin Ma$^{\ddagger}$\renewcommand{\thefootnote}{\fnsymbol{footnote}}\footnotemark[4]\qquad Wei Liu$^{\ddagger}$\qquad Yong Xu$^{\dagger}$\\
$^{\dagger}$South China University of Technology \qquad $^{\ddagger}$Tencent AI Lab
\\
{\tt\small$\lbrace$jaywongjaywong, cswhjiang, forest.linma$\rbrace$@gmail.com} \\
{\tt\small wliu@ee.columbia.edu\qquad yxu@scut.edu.cn}
}

\maketitle
\renewcommand{\thefootnote}{\fnsymbol{footnote}}
\footnotetext[4]{Corresponding authors.}

\begin{abstract}
Dense video captioning is a newly emerging task that aims at both localizing and describing all events in a video. We identify and tackle two challenges on this task, namely, (1) how to utilize both past and future contexts for accurate event proposal predictions, and (2) how to construct informative input to the decoder for generating natural event descriptions. First, previous works predominantly generate temporal event proposals in the forward direction, which neglects future video context. We propose a bidirectional proposal method that effectively exploits both past and future contexts to make proposal predictions. Second, different events ending at (nearly) the same time are indistinguishable in the previous works, resulting in the same captions. We solve this problem by representing each event with an attentive fusion of hidden states from the proposal module and video contents (e.g., C3D features). We further propose a novel context gating mechanism to balance the contributions from the current event and its surrounding contexts dynamically. We empirically show that our attentively fused event representation is superior to the proposal hidden states or video contents alone. By coupling proposal and captioning modules into one unified framework, our model outperforms the state-of-the-arts on the ActivityNet Captions dataset with a relative gain of over \textbf{100\%} (Meteor score increases from \textbf{4.82} to \textbf{9.65}).
\end{abstract}

\section{Introduction}
With the rapid growing of videos on the Internet, it becomes much more important to automatically classify and retrieve these videos. While images and short videos have attracted extensive attentions from vision research community \cite{xu2009viewpoint,linma2015iccv,wang2016beyond,ren2016single,wenhaojiang2018aaai,DevNet,ren2017video,chen2016sca,linma2016aaai}, understanding long untrimmed videos remains an open  question. To help further understand videos and bridge them with human language, a new task of dense video captioning is proposed \cite{Krishna_2017_ICCV}. The goal is to automatically localize events in videos and describe each one with a sentence. The capability of localizing and describing events in videos will benefit a broad range of applications, such as video summarization \cite{lu2013story,potapov2014category}, video retrieval \cite{snoek2008concept,yang2014content}, video object detection \cite{Yuan_2017_ICCV}, video segment localization with language query \cite{Hendricks_2017_ICCV,gao2017tall}, and so on.

\begin{figure}[t]
\begin{center}
\includegraphics [width=1.0\linewidth]{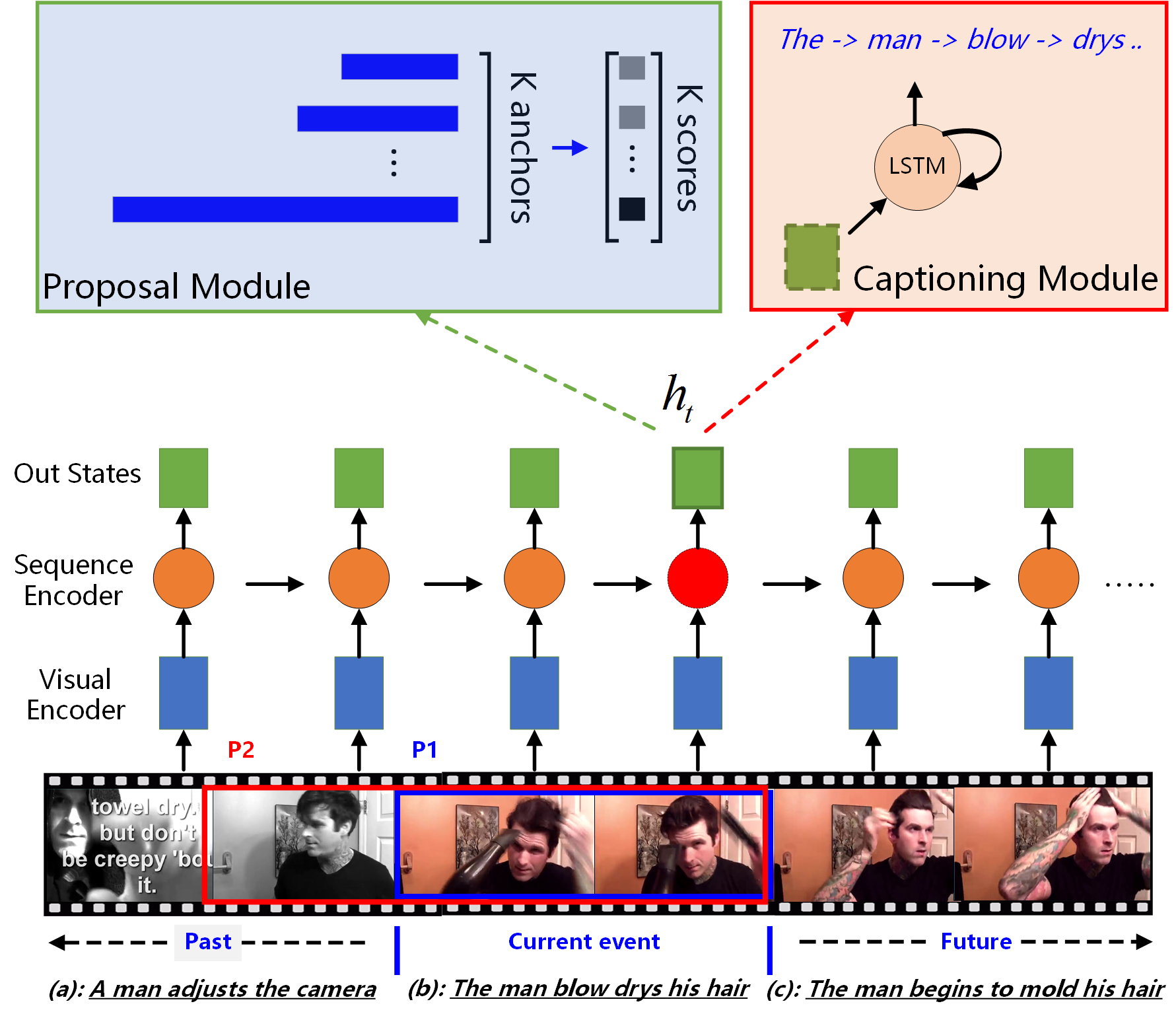}
\caption{Two main challenges in dense video captioning. First, previous works, \textit{e.g.}, SST \cite{Buch_2017_CVPR}, process a video sequence in the forward direction. Future video context (c) expressing ``\texttt{\small The man begins to mold his hair}" is not considered, which presents close relationship with current proposal (b) expressing ``\texttt{\small The man blow drys his hair}''. Second, previous work only uses the proposal hidden state $h_t$ at time step $t$ to represent the detected proposal, which cannot distinguish events (\textit{e.g.}, P1, P2) that end at the same time step.}
\vspace{-0.3cm}
\label{figure_introduction}
\end{center}
\end{figure}

Compared to video captioning, which targets at describing a short video clip (e.g., 20s long in MSR-VTT dataset \cite{xu2016msr}), dense video captioning requires analyzing a much longer and complicated video sequence (e.g., 120s long in ActivityNet Captions \cite{Krishna_2017_ICCV}). Since long videos usually involve multiple events, dense video captioning requires simultaneously performing temporal localization and captioning, which issues the following two challenges.

First, generating video action proposals requires localizing all possible events that occur in a video. To do so, one simple way would be to use sliding windows to iterate over a video and classify every window to either an action or background. However, this kind of methods can only produce short proposals that are no longer than the predefined sliding window. To overcome this problem, Buch \emph{et al.} \cite{Buch_2017_CVPR} proposes Single Stream Temporal Action Proposals (SST) to eliminate the need to divide long video sequences into clips or overlapped temporal windows. As shown in Fig. \ref{figure_introduction}, SST runs through a video only once and densely makes proposal predictions ending at that time step, with \emph{k} different offsets. Krishna \emph{et al.} \cite{Krishna_2017_ICCV} uses a similar proposal method as SST. While promising results were achieved, these methods simply ignore future event context and only encode past context and current event information to make predictions. Since events happening in a video are usually highly correlated, it is non-preferable to discard valuable future information. For example, in Fig. \ref{figure_introduction}, when making proposal prediction at the end of event (b), SST has run over both past context (a) and current event content, but not future video context (c). Event (b) highly correlates with event (c). Recognizing and localizing event (b) will help localize event (c) more accurately, and vice versa. In this paper, we propose a straightforward yet effective solution, namely, Bidirectional SST, towards efficiently encoding both past, current, and future video information. Specifically, in the forward pass we learn \emph{k} binary classifiers corresponding to \emph{k} anchors densely at each time step; in backward pass we reverse both video sequence input and predict proposals backwards. This means that the forward pass encodes past context and current event information, while the backward pass encodes future context and current event information. Finally we merge proposal scores for the same predictions from the two passes and output final proposals. Technical details can be found in Section \ref{method}.

Once proposals are obtained, another important question is how to represent these proposals in order to generate language descriptions. In \cite{Krishna_2017_ICCV}, the  LSTM hidden state in proposal module is reused to represent a detected proposal. However, the discrimination property of event representation is overlooked. As shown in Fig. \ref{figure_introduction}, \emph{k} proposals (anchors) end at same time step, but only one LSTM hidden state $h_t$ at that time step is returned. For example, $P1$ and $P2$ will be both represented with $h_t$. To construct more discriminative proposal representation, we propose to fuse proposal state information and detected video content (e.g. C3D sequences). The intuition behind that is involving detected clips help discriminate highly overlapped events, since the detected temporal regions are different. Based on this idea, we further explore several ways for fusing these two kinds of information to boost dense captioning performance.

To output more confident results, we further propose joint ranking technique to select high-confidence proposal-caption pairs by taking both proposal score and caption confidence into consideration.

To summarize, the contributions of this paper are three-fold. First, we present Bidirectional SST for better temporal action proposals with both past, current, and future contexts encoded. Second, for captioning module, we explore different ways to attentively fuse proposal state information and detected video content to effectively discriminate highly overlapped events. Third, we further present joint ranking at inference time to select proposal-caption pairs with high confidence.

\section{Related Work}
\label{related_work}
Dense video captioning requires both temporally localization and descriptions for all events happening in a video. These two tasks can be handled as pipelines or coupled together for end-to-end processing. We review related works on the above two tasks.

\subsection{Temporal Action Proposals}
Analogous to region proposals in image domain, temporal action proposals are candidate temporal windows that possibly contain actions. Sparse-prop~\cite{caba2016fast} applies dictionary learning for generating class-independent proposals. S-CNN \cite{shou2016temporal} uses 3D convolutional neural networks  (CNNs)~\cite{tran2015learning} to generate multi-scale segments (proposals). TURN~TAP~\cite{Gao_2017_ICCV} uses clip pyramid features in their model, and it predicts proposals and refines temporal boundaries jointly. DAPs \cite{escorcia2016daps} first applies Long Short-Term Memory (LSTM) \cite{hochreiter1997long} to encoding video content in a sliding window and then predicts proposals covered by the window. Built on \cite{escorcia2016daps}, SST \cite{Buch_2017_CVPR} further takes long sequence training problem into consideration and generates proposals in a single pass. However, all these methods either fail to produce long proposals or do not exploit future context. In contrast, our model for temporal proposal tackles these two problems simultaneously.

\begin{figure*}
\centering
\includegraphics [width=0.95\linewidth]{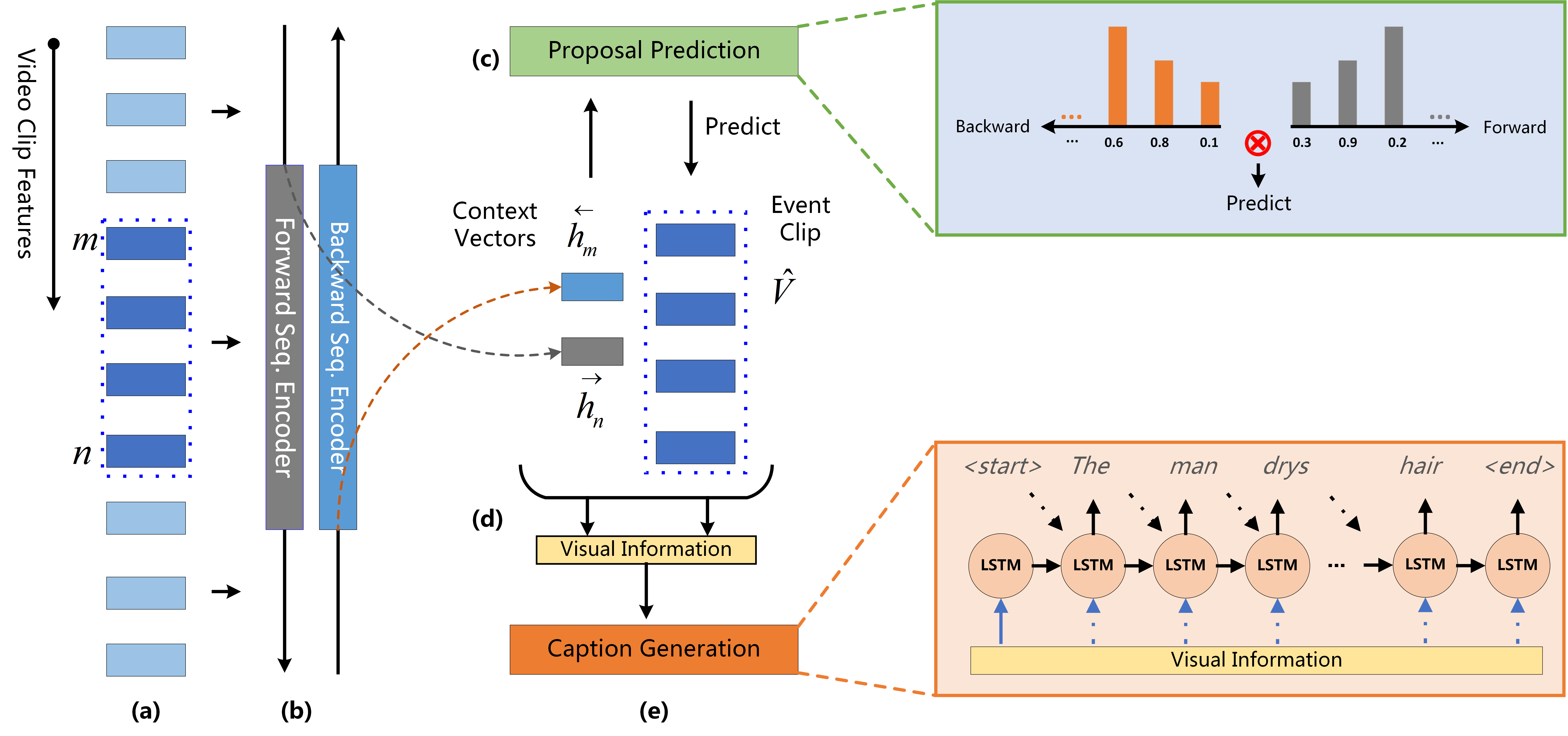}
\caption{The main framework of our proposed method. (a) A video input is first encoded as a sequence of visual features (e.g., C3D). (b) The visual features are then fed into our bidirectional sequence encoder (e.g., LSTM). (c) Each hidden state from the forward/backward seq. encoder will be fed into the proposal module. The forward/backward seq. encoders are jointly learned to make  proposal predictions. (d) Hidden states at the boundary of a detected event ($\overset{\rightarrow}{\mathbf{h}_n}$, $\overset{\leftarrow}{\mathbf{h}_m}$) will be served as context vectors for the event. The context vectors and detected event clip features are then fused together and served as visual information input. We detail the fusion methods in Section \ref{subsubsection_fusion}. (e) The decoder LSTM translates visual input into a sentence.}
\label{figure_method}
\vspace{-0.3cm}
\end{figure*}

\subsection{Video Captioning} 

\noindent\textbf{Video captioning with one single sentence.} There are a large body of works on this topic. Earlier works are template-based \cite{guadarrama2013youtube2text,rohrbach2013translating}, which replace POS (part-of-speech) tags with detected objects, attributes, and places. \cite{guadarrama2013youtube2text} learns semantic hierarchies from video data in order to choose an appropriate level of sentence descriptions. \cite{rohrbach2013translating} first formulates video captioning as a machine translation problem and uses CRF to model semantic relationship between visual components. Recent approaches are neural-based, in an encoder-decoder fashion \cite{kiros2014multimodal,venugopalan2015sequence,pan2016jointly,yao2015describing,Zhang_2017_CVPR,Pan_2017_CVPR,Gan_2017_CVPR,chen2017generating,long2016video}. Venugopalan \emph{et al.} models both video and language as sequences using recurrent neural networks \cite{venugopalan2015sequence}. To strengthen the semantic relationships between a video and the corresponding generated sentence, Pan \emph{et al.} proposed to  learn the translation and a common embedding space shared by video and language jointly~\cite{pan2016jointly}. Some subsequent methods further explore attention models in video context. Inspired by the soft attention mechanism~\cite{xu2015show} in image captioning, Yao \emph{et al.} proposed to generate temporal attention over video frames when predicting next word~\cite{yao2015describing}. Zhang \emph{et al.} proposed to learn a task-driven fusion model by dynamically fusing complementary features from multiple channels (appearance, motion) \cite{Zhang_2017_CVPR}. Some other works \cite{Pan_2017_CVPR,Gan_2017_CVPR,Yao_2017_ICCV} exploit attributes or concepts (objects, actions, etc.) to improve video captioning performance. \cite{chen2017generating} further considers different topics from web videos and generates topic-guided descriptions.

\vspace{5pt}
\noindent\textbf{Video captioning with a paragraph.} While aforementioned captioning methods generate only one sentence for an input video, video paragraph generation focuses on producing multiple semantics-fluent sentences. Rohrbach \emph{et al.} adapted statistical machine translation (SMT) \cite{rohrbach2013translating} to generate semantic consistent sentences with desired level of details \cite{rohrbach2014coherent}. Yu \emph{et al.} proposed a hierarchical RNN to model both cross-sentence dependency and word dependency \cite{yu2016video}.

\vspace{5pt}
\noindent\textbf{Dense video captioning.} Video paragraph generation relies on alignment from \emph{ground-truth event intervals} at test time. To relieve this constraint, dense video captioning generates multiple sentences and grounds them with time locations automatically, which is thus much more challenging. To the best of our knowledge, \cite{Krishna_2017_ICCV} is the only published work on this topic. In~\cite{Krishna_2017_ICCV}, the task of dense-captioning events in video together with a new dataset: ActivityNet Captions\renewcommand{\thefootnote}{\arabic{footnote}}\footnote{http://cs.stanford.edu/people/ranjaykrishna/densevid/} were introduced. The model in~\cite{Krishna_2017_ICCV} is composed of an event proposal module and a captioning module. The event proposal module detects events with a multi-scale version of DAPs~\cite{escorcia2016daps} and represents them with LSTM hidden states. The captioning module is responsible for describing each detected proposal. Compared to \cite{Krishna_2017_ICCV}, our method enjoys the following advantages. First, our bidirectional proposal module encodes both past and future contexts while \cite{Krishna_2017_ICCV} only utilizes past context for proposal prediction. Second, our model is able to distinguish and describe highly overlapped events while \cite{Krishna_2017_ICCV} cannot. 

\section{Method}
\label{method}
In this section we introduce our main framework for densely describing events in videos, as shown in Fig. \ref{figure_method}. We will first introduce our bidirectional proposal module, then our captioning module. Note that these two modules couple together and thus can be trained in an end-to-end manner.

\subsection{Proposal Module}
\label{proposal_module}
The goal of the proposal module is to generate a set of temporal regions that possibly contain actions or events. Formally, assume that we have a video sequence ${X}=\{\mathit{x}_1, \mathit{x}_2, ..., \mathit{x}_L\}$ with $\mathit{L}$ frames. Following \cite{Krishna_2017_ICCV}, each video frame is encoded by  the 3D CNN \cite{tran2015learning}, which was pre-trained on Sports-1M video dataset \cite{karpathy2014large}. The extracted C3D features are of temporal resolution $\mathit{\delta}$ = 16 frames, discretizing the input stream into $\mathit{T=L/\delta}$ time steps. We perform PCA to reduce the feature dimensionality (from 4096 to 500). The generated visual stream is thus $\mathit{V}=\{\mathbf{v}_1, \mathbf{v}_2,..., \mathbf{v}_T\}$. 

\vspace{5pt}
\noindent\textbf{Forward Pass.} We use LSTM to sequentially encode the visual stream. The sequence encoder processes visual sequences and accumulates visual clues across time. The output LSTM hidden state $\mathit{\overset{\rightarrow}{\mathbf{h}_t}\in\big\{\overset{\rightarrow}{\mathbf{h}_i}\big\}_{i=1}^T}$ at time step $\mathit{t}$ thus encodes visual information for the passed time steps $\{1, 2, \dots, \mathit{t}\}$. The hidden state will be fed into ${K}$ independent binary classifiers and produces ${K}$ confidence scores $\mathit{\overset{\rightarrow}{C_p}^{t}=\big\{\overset{\rightarrow}{c_i}^{t}\big\}}_{i=1,...,{K}}$ indicating the probabilities of $K$ proposals specified by $\mathit{\overset{\rightarrow}{S}^{t}=\big\{\overset{\rightarrow}{s_i}^{t}\big\}}_{i=1,...,{K}}$. $\mathit{\overset{\rightarrow}{s_i}^{t}}$ denotes a video clip with end time as $t$ and start time as $t-\mathit{l}_i$, where $\{\mathit{l}_i\}_{i=1}^K$ is the lengths of the predefined $K$ proposal anchors. Please note that all the ${K}$ proposals in $\overset{\rightarrow}{S}^{t}$ share the same end time $t$. The proposal scores $\overset{\rightarrow}{C_p}^{t}$ are calculated by a fully connected layer:
\begin{align}
\label{equation_score_sigmoid}
\overset{\rightarrow}{C_p}^{t} = \sigma(\overset{\rightarrow}{W_c} \overset{\rightarrow}{\mathbf{h}_t} + \mathbf{b}_c),
\end{align}
where $\sigma$ denotes \emph{sigmoid} nonlinearity. $\overset{\rightarrow}{W_c}$, $\mathbf{b}_c$ are shared across all time steps.

\vspace{5pt}
\noindent\textbf{Backward Pass.} Our proposed bidirectional proposal module also involves a backward pass. The aim of such a procedure is to capture future context, in addition to current event clue for better event proposals. We feed the input sequence $\mathit{V}$ in a reverse order to the backward sequence encoder. It is expected to predict proposals with high scores at the original start time of proposals. Similarly, at each time step, we obtain $K$ proposals $\mathit{\overset{\leftarrow}{S}^{t}=\big\{\overset{\leftarrow}{s_i}^{t}\big\}}_{i=1,...,K}$ with $K$ confidence scores $\mathit{\overset{\leftarrow}{C_p}^{t}=\big\{\overset{\leftarrow}{c_i}^{t}\big\}}_{i=1,...,K}$, and a hidden state $\mathit{\overset{\leftarrow}{\mathbf{h}_t}}$.

\vspace{5pt}
\noindent\textbf{Fusion.} After the two passes, we obtain $\mathit{N}$ proposals collected from all time steps of both directions.
In order to select proposals with high confidence, we fuse the two sets of scores for the same proposals, yielding the final scores:
\begin{equation}
\label{equation_combine_score_bidirection}
C_p=\big\{\overset{\rightarrow}{c_i} \times \overset{\leftarrow}{c_i}\big\}_{i=1}^N.
\end{equation}
Many fusing strategies can be adopted. In this paper, we simply use the multiplication to fuse proposals from the two passes together. Proposals with scores larger than a threshold $\mathit{\tau}$ will be finally selected for further captioning. We do not perform non-maximum suppression since events happening in a video are usually highly overlapped, the same as what has been adopted in \cite{Krishna_2017_ICCV}.
\subsection{Captioning Module}
\label{caption_module}
Following the encoder-decoder framework, a recurrent neural network, specifically LSTM, is leveraged in our captioning module to translate visual input into a sentence. In this section, we first recap LSTM. Then we describe a novel dynamic fusion method. 

\subsubsection{Decoder: Long Short-Term Memory}
LSTM~\cite{hochreiter1997long} is used as our basic building block, considering its excellent ability for modeling sequences. An LSTM unit consists of an input cell $\mathit{g_t}$, an input gate $\mathit{i_t}$, a forget gate $\mathit{f_t}$, and an output gate $\mathit{o_t}$ and they can be computed by:
\begin{equation}
\label{equation_lstm}
\begin{split}
  \begin{pmatrix} i_t \\ f_t \\ o_t \\ g_t \end{pmatrix} &=
  \begin{pmatrix} \sigma \\ \sigma \\ \sigma \\ \tanh \end{pmatrix}
 W
  \begin{pmatrix} \mathbf{E}_t \\ \mathbf{F}_t \\ \mathbf{H}_{t-1} \end{pmatrix}\\
\end{split},
\end{equation}
where $\mathbf{E}_t$ is the embedding of input word at time step $t$, $\mathbf{F}_t$ is representation at $t$ that will be described later, $\mathbf{H}_{t-1}$ is the previous LSTM hidden state and $W$ is a transformation matrix to be learned. The memory cell $c_t$ and hidden state  $\mathbf{H}_{t}$ are updated by:
\begin{align}
c_t &=f_t \odot c_{t-1}+i_t \odot g_t,\\
\mathbf{H}_t &=o_t \odot \tanh(c_t),
\end{align}
where $\odot$ denotes element-wise multiplication operator. At each time step, a linear projection and softmax operation are performed on the hidden state to generate probability distribution over all possible words.

\subsubsection{Dynamic Attentive Fusion with Context Gating}
\label{subsubsection_fusion}
To caption a detected proposal, previous work just takes the proposal hidden state as input to the LSTM \cite{Krishna_2017_ICCV}. In this paper we propose to fuse the proposal states from the forward and backward passes, which capture both past and future contexts, together with the encoded visual features of the detected proposal. Formally, the visual input to the decoder is:
\begin{equation}
\label{equation_proposal_representation}
\mathbf{F}_t(s_i)=f(\overset{\rightarrow}{\mathbf{h}_n}, \overset{\leftarrow}{\mathbf{h}_m}, \hat{V}=\{\mathbf{v}_i\}_{i=m}^n, \mathbf{H}_{t-1}),
\end{equation}
where $\mathit{m}$ and $\mathit{n}$ denote the start and end time stamp for the detected event $\mathit{s_i}$. $\hat{V}$ denotes the clip features, specifically C3D for the proposal $s_i$. 
$\overset{\rightarrow}{\mathbf{h}_n}$ and $\overset{\leftarrow}{\mathbf{h}_m}$ are the proposal hidden states, encoding the past and future context information of the detected proposal, which are simply named context vectors. $\mathbf{H}_{t-1}$ is the previous LSTM hidden state. And $f$ is a mapping to output a compact vector, which is to be fed into LSTM unit using Eq.~\eqref{equation_lstm}.

The most straightforward way is to simply concatenate $\mathit{\hat{V}}$, $\overset{\rightarrow}{\mathbf{h}_n}$, and $\overset{\leftarrow}{\mathbf{h}_m}$ together without considering $\mathbf{H}_{t-1}$. However, it is implausible, as the dimension of $\mathit{\hat{V}}$ depends on the length of a detected event. Another simple way is to use the mean of $\mathit{\hat{V}}$ and concatenate it with proposal hidden states. However, mean pooling does not explicitly explore relationship between an event and surrounding contexts.
\vspace{5pt}
\noindent\textbf{Temporal Dynamic Attention.} As demonstrated in ~\cite{xu2015show,yao2015describing}, dynamically attending on image sub-regions and video frames at each time step when decoding can effectively improve captioning performance. Therefore, in our dense captioning model, we also design a dynamic attention mechanism to fuse visual features $\mathit{\hat{V}}$ and context vectors $\overset{\rightarrow}{\mathbf{h}_n}$, $\overset{\leftarrow}{\mathbf{h}_m}$. At each time step $t$, the  relevance score $z_i^{t}$ for $\mathbf{v}_{i+m-1}$ is obtained by:
\begin{equation}
\label{equation_relevance_score}
z_i^{t} = W_a^T \cdot \tanh (W_\mathbf{v}\mathbf{v}_{i+m-1} + W_h[{\overset{\rightarrow}{\mathbf{h}_n}}, {\overset{\leftarrow}{\mathbf{h}_m}}] + W_\mathbf{H}\mathbf{H}_{t-1} + \mathbf{b}),
\end{equation}
where $\mathbf{H}_{t-1}$ is the hidden states of decoder at the $t-1$ time step. $[\cdot,\cdot]$ denotes vector concatenation. The weights of $\mathbf{v}_{i+m-1}$ can be obtained by a softmax normalization:
\begin{equation}
\label{equation_attention}
\alpha_i^{t} = \exp(z_i^{t})/\sum_{k=1}^p \exp(z_k^{t}),
\end{equation}
where $p=n-m+1$ denotes the length of a proposal. The attended visual feature is generated by a weighted sum:
\begin{equation}
\tilde{\mathbf{v}}^{t}=\sum_{i=1}^p \alpha_{i}^{t} \cdot \mathbf{v}_{i+m-1}.
\end{equation}
We expect the model can better locate ``key frames'' and produce more semantic correlated words by involving context vectors for calculating the attention as in Eq. \eqref{equation_relevance_score}. The final input to LSTM unit could be expressed as:
\begin{equation}
\mathbf{F}(s_i) = [\tilde{\mathbf{v}}^t, \overset{\rightarrow}{\mathbf{h}_n}, \overset{\leftarrow}{\mathbf{h}_m}].
\end{equation}

\vspace{5pt}
\noindent\textbf{Context Gating.} Inspired by the gating mechanism in LSTM, we propose to explicitly model the relative contributions of the attentive event feature and contexts when generating a word. Specifically, once obtain the attended visual feature $\tilde{\mathbf{v}}^{t}$, instead of directly concatenating it with context vectors, we learn a ``context gate'' to balance them. In our context gating mechanism, the first step is to project the event feature and the context vectors into the same space:
\begin{equation}
\dot{\mathbf{v}}^{t} = \tanh (\tilde{W} \tilde{\mathbf{v}}^{t}), 
\end{equation}
\begin{equation}
\mathbf{h} = \tanh (W_{ctx} [ \overset{\rightarrow}{\mathbf{h}_n}, \overset{\leftarrow}{\mathbf{h}_m} ]),
\end{equation}
where $\tilde{W}$ and $W_{ctx}$ are the projection matrices. The context gate is then calculated by a nonlinear layer:
\begin{equation}
g_{ctx} = \sigma (W_g [\dot{\mathbf{v}}^{t}, \mathbf{h}, \mathbf{E}_t, \mathbf{H}_{t-1}]),
\end{equation}
where $\mathbf{E}_t$ is word embedding vector, $\mathbf{H}_{t-1}$ is the previous LSTM state. The context gate explicitly measures the contribution for the surrounding context information ($\mathbf{h}$) at current decoding stage (given $\mathbf{E}_t$, $\mathbf{H}_{t-1}$). We then use the context gate to fuse the event feature and the context vector together:
\begin{equation}
\mathbf{F}(s_i) = [(1-g_{ctx}) \odot \dot{\mathbf{v}}^{t}, g_{ctx} \odot \mathbf{h}].
\end{equation}
With this mechanism, we expect the network to learn how much context should be used when generating next word.

\subsection{Training}
Our complete dense video captioning model, as illustrated in Fig. \ref{figure_method}, couples the proposal and captioning module together. Therefore, two types of loss functions are considered in our model, specifically, the proposal loss and captioning loss.

\vspace{5pt}
\noindent\textbf{Proposal Loss.}
We collect lengths of all ground-truth proposals and group them into $K$=128 clusters (anchors). Each training example $V=\{\mathbf{v}_i\}_{i=1}^T$ is associated with ground-truth labels $\{y_t\}_{t=1}^T$. Each $y_t$ is a $K$-dim vectors with binary entries.  $y_t^j$ is set to 1 if the corresponding proposal interval has a temporal Intersection-over-Union (tIoU) with the ground-truth larger than 0.5 and set to 0 otherwise. We adopt weighted multi-label cross entropy as proposal loss $\mathcal{L}_{p}$ following ~\cite{Buch_2017_CVPR} to balance positive and negative proposals. For a given video $X \in \mathcal{X}$ at time step $t$:
\begin{equation}
\mathcal{L}_{p}(c,t,X,y) = -\sum_{j=1}^K w_0^j y_t^j log c_t^j + w_1^j (1-y_t^j) log (1-c_t^j),
\end{equation}
where $w_0^j$, $w_1^j$ are determined based on the numbers of positive and negative proposal samples. $c_t^j$ is the prediction score for the $j$-th proposal at time $t$. We calculate forward and backward loss in the same way. We add them together and jointly train the forward and backward proposal module. $\mathcal{L}_{p}$ is obtained by averaging along time steps and for all videos.

\vspace{5pt}
\noindent\textbf{Captioning Loss.} We only feed proposals of high tIoU ($>$ 0.8) with ground-truths to train captioning module. Following ~\cite{vinyals2015show}, we define captioning loss $\mathcal{L}_{c}$ as sum of negative log likelihood of \emph{correct word} in a sentence with $M$ words:
\begin{equation}
\mathcal{L}_{c}(P) = -\sum_{i=1}^M \log(p(w_i)),
\end{equation}
where $w_i$ is the $i$-th word in a ground truth sentence. $\mathcal{L}_{c}$ is obtained by averaging all $\mathcal{L}_{c}(P)$ for all proposals $P$.

\vspace{5pt}
\noindent\textbf{Total Loss.} By considering both proposal localization and captioning, the total loss is given by:
\begin{equation}
\mathcal{L} = \lambda \times \mathcal{L}_{p} + \mathcal{L}_{c},
\end{equation}
where $\lambda$ balances the contributions between proposal localization and captioning, which is simply set to 0.5. A two-layer LSTM is used to encode a video stream and it densely predicts $K$ proposals at each time step. For fair comparison, we initialize our bidirectional sequence encoder with a single layer LSTM for each direction (the baseline method adopts a two-layer LSTM). We use a two-layer LSTM during decoding stage (caption generation).

\subsection{Inference by Joint Ranking}
As illustrated in Fig. \ref{figure_method}, dense captioning involves the two aforementioned modules. As such, to affectively describe each event, two conditions need to be satisfied: (1) the localization yielded by proposal module is of high score; (2) the produced caption is of high confidence. To this end, we propose a novel joint ranking approach for dense captioning during the inference stage. We use Eq.~\eqref{equation_combine_score_bidirection} to measure the proposal score $C_p$. For a generated caption of a proposal consisting of $M$ words $\{w_i\}_{i=1}^M$, we define its confidence by summing all log probabilities of predicted words:
\begin{equation}
c_c = \sum_{i=1}^M \log(p(w_i)).
\end{equation}
Larger $p(w_i)$ indicates higher confidence score. Let $C_c=\big\{c_c^{(i)}\big\}_{i=1}^N$ denotes confidence scores of all sentences. We merge the two scores with a weighted sum strategy by simultaneously considering proposal localization and captioning:
\begin{equation}
C = \gamma\times C_p + C_c,
\end{equation}
where $\gamma$ is a trade-off parameter to control the contributions from localization and captioning. As $C_p$ is of smaller scale, $\gamma$ is empirically set as 10 in this paper. Based on the obtained $C$, Top ${K}$ proposals together with their captions are selected for further evaluation.

\section{Experiment}
To detail our contributions, we conduct experiments on the following tasks: (1) event localization, (2) video captioning with ground truth proposals, and (3) dense video captioning. The first evaluates how good the generated proposals are, the second measures the quality of our captioning module, and the third measures the performance of our whole dense captioning system.

\vspace{5pt}
\noindent \textbf{Datasets.} (1) ActivityNet Captions~\cite{Krishna_2017_ICCV} is built on ActivityNet v1.3~\cite{caba2015activitynet} which includes 20k YouTube untrimmed videos from real life. The videos are 120 seconds long on average. Most of the videos contain over 3 annotated events with corresponding start/end time and human-written sentences, which contain 13.5 words on average. The number of videos in train/validation/test split is 10024/4926/5044, respectively. Ground truth annotations from the test split are withheld for competition. We use this dataset for all the three tasks. We first compare our model with baseline methods on validation set, then we report our final result returned from the test server. (2) THUMOS-14~\cite{jiang2014thumos} has 200 videos for training and 213 videos for testing. The videos are 200 seconds long on average. Each video is associated with multiple action proposals, annotated with their action labels and time boundaries. We use this dataset to evaluate different proposal methods. We follow the experimental setting from \cite{Buch_2017_CVPR}.

\begin{figure*} [!htb]
	\begin{center}
		\begin{tabular}{@{}ccc@{}}
			\includegraphics[width = 0.3\textwidth]{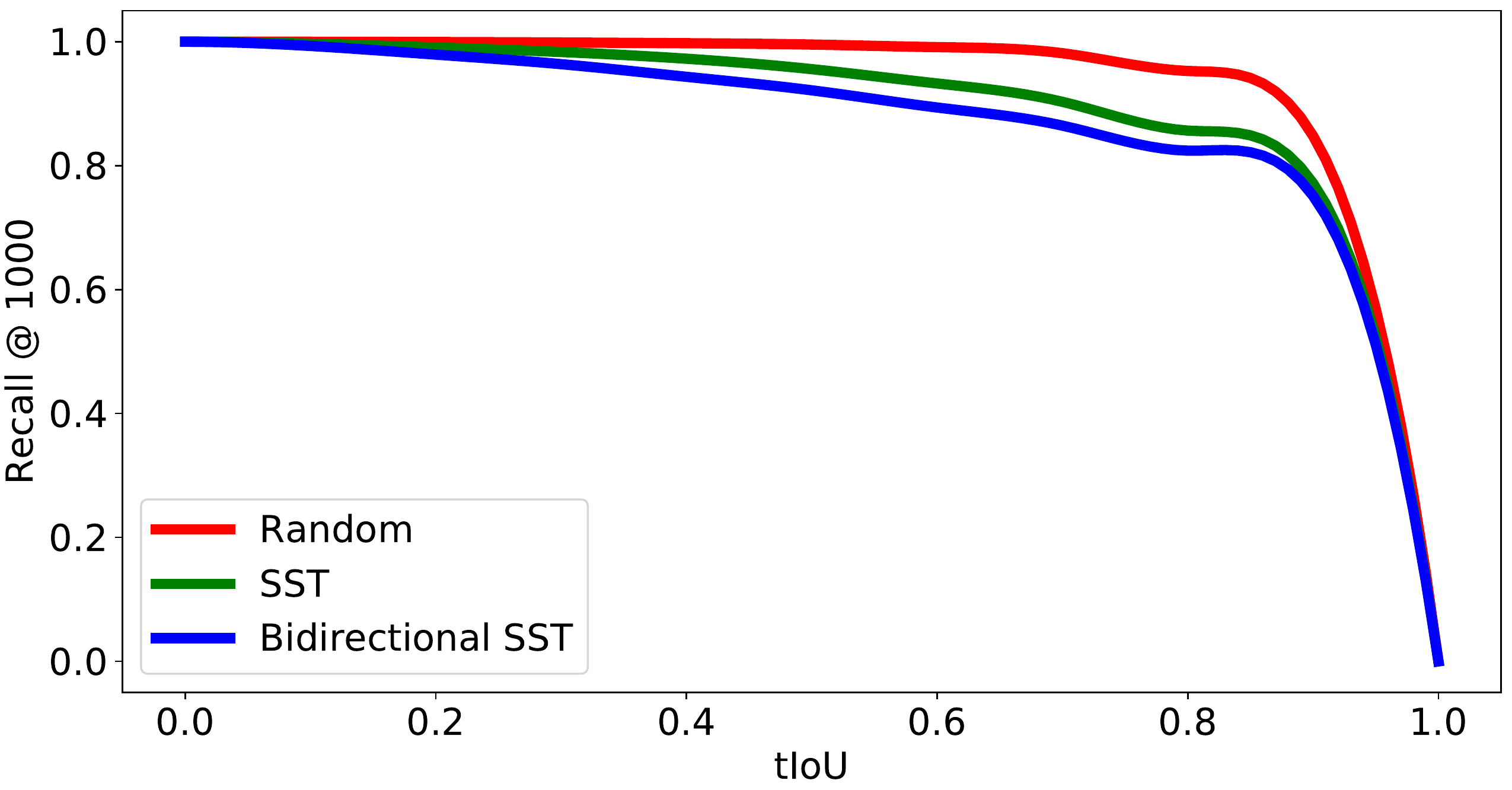} & \includegraphics[width = 0.3\textwidth]{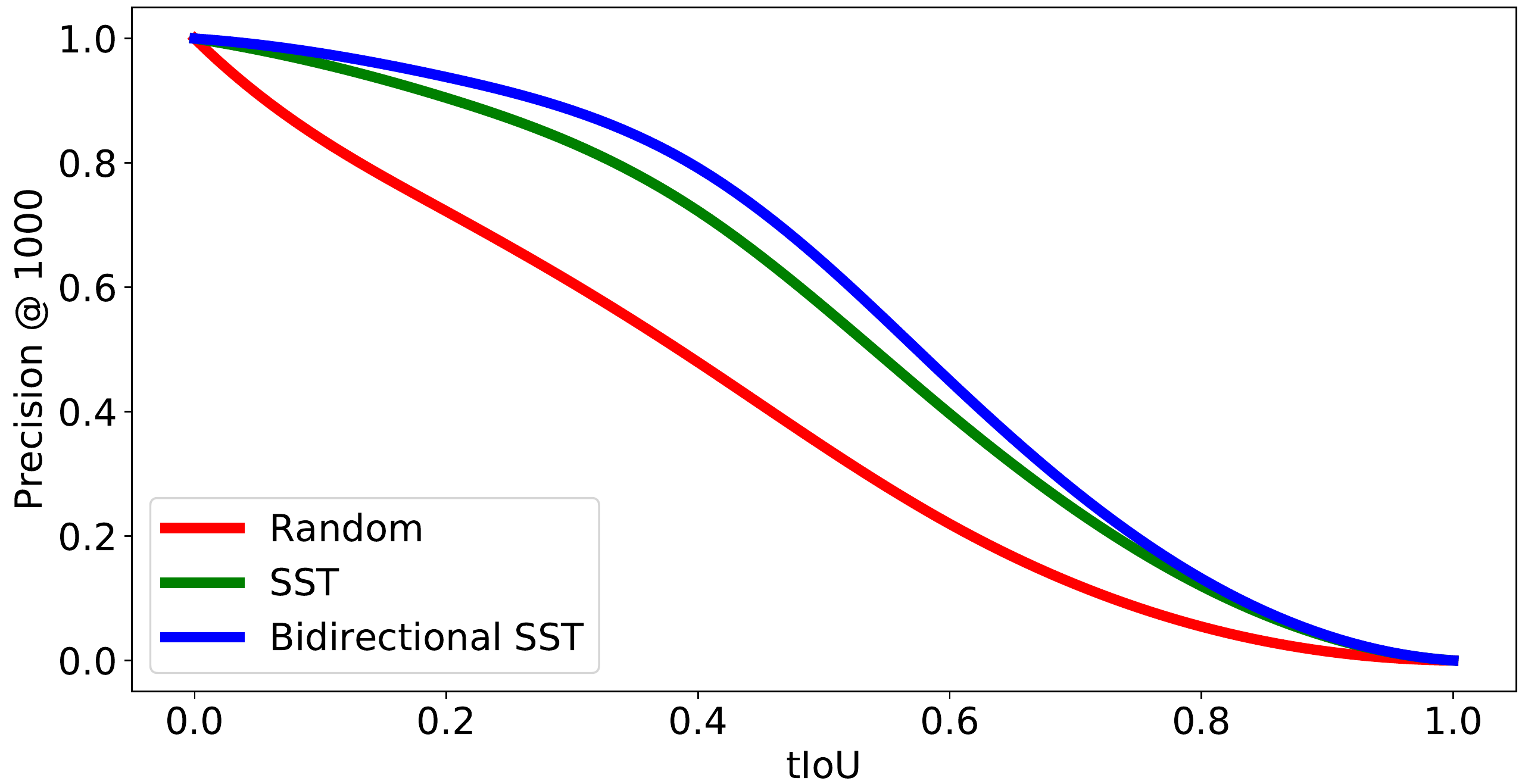} & \includegraphics[width = 0.3 \textwidth]{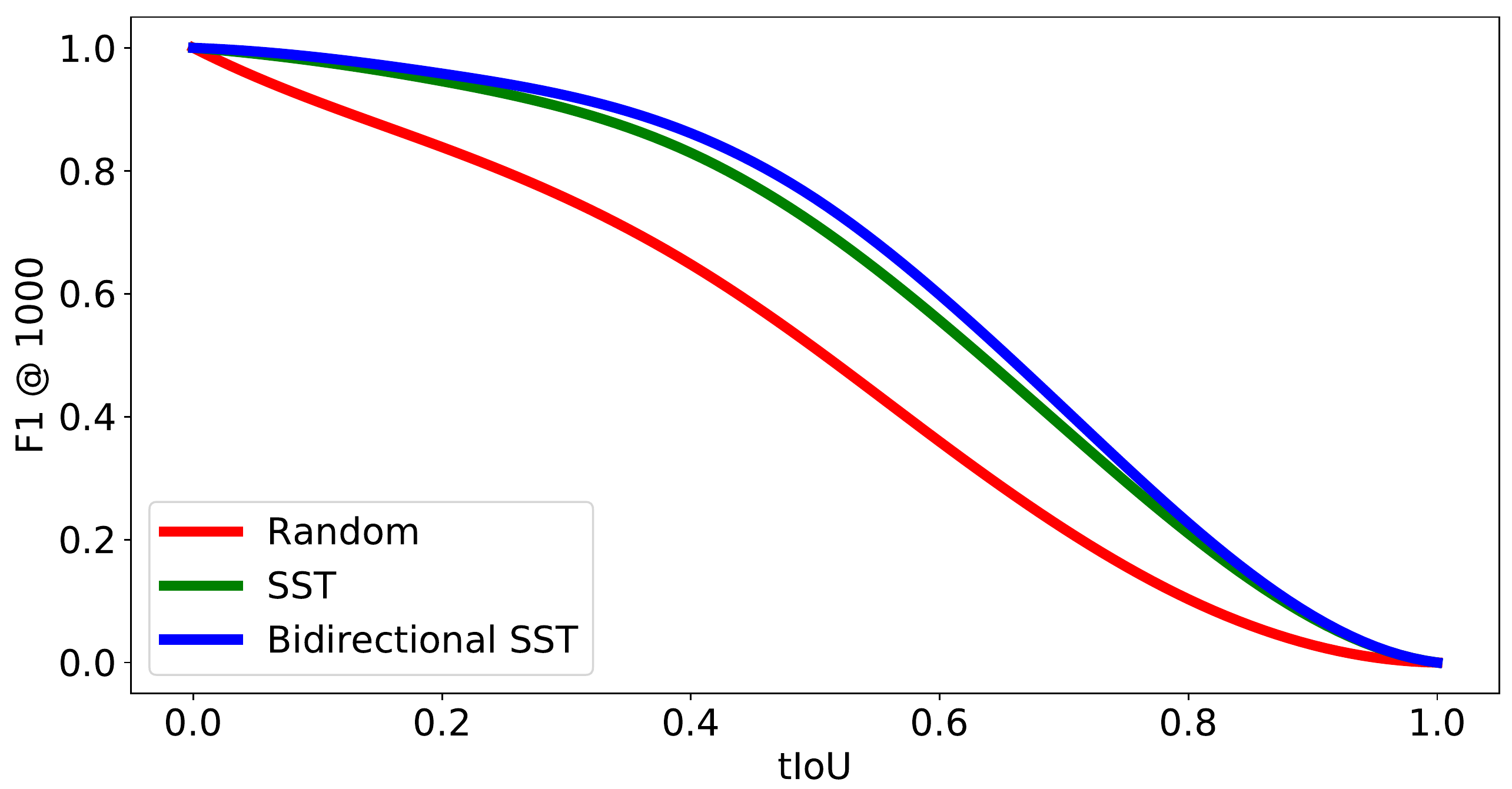}\\
			(a) Recall@1000 \emph{vs.} tIoU & (b) Precision@1000 \emph{vs.} tIoU & (c) F1@1000 \emph{vs.} tIoU
		\end{tabular}
	\end{center}
    \vspace{-5pt}
	\caption{Comparison for different proposal methods.}
	\label{figure_precision_recall_f1}
\end{figure*}

\begin{table}
{\footnotesize
\caption{Comparison of SST and Bidirectional SST on ActivityNet Captions. Our Bidirectional SST surpasses the Random baseline and Forward/Backward SST with clear margins.}
\vspace{-5pt}
\label{table_event_localization}
\begin{center}
\begin{tabular} {c|c|c|c}
\hline
\textbf{Method} & \textbf{Pre@1000} & \textbf{Rec@1000} & \textbf{F1@1000}\\
\hline\hline
Random & 0.272 & \textbf{0.956} & 0.424\\
\hline
Forward SST & 0.411 & 0.910 & 0.566 \\
\hline
Backward SST & 0.441 & 0.856 & 0.582 \\
\hline
Bidirectional SST & \textbf{0.459} & 0.875 & \textbf{0.602} \\
\hline
\end{tabular}
\end{center}
}
\vspace{-10pt}
\end{table}

\begin{table}
{\footnotesize
\caption{Recall@\textit{k} proposals of SST and Bi-SST on THUMOS-14~\cite{jiang2014thumos} dataset. The results are averaged on tIoUs of 0.5 to 1.0 as~\cite{escorcia2016daps}.}
\vspace{-7pt}
\label{table_proposal_avg_tiou}
\begin{center}
\begin{tabular} {c|c|c|c|c}
\hline
\textbf{Method} & \textbf{@10} & \textbf{@100} & \textbf{@200} & \textbf{@1000}\\
\hline\hline
SST (our impl) & 0.053 & 0.259 & 0.372 & 0.628 \\
\hline
Backward SST & 0.059 & 0.273 & 0.386 & 0.614 \\
\hline
Bi-SST & \textbf{0.063} & \textbf{0.285} & \textbf{0.393} & \textbf{0.633} \\
\hline
\end{tabular}
\end{center}
}
\vspace{-10pt}
\end{table}

\begin{table}
{\footnotesize
\caption{Recall@1000 proposals at tIoU of 0.8 of SST and Bi-SST on THUMOS-14~\cite{jiang2014thumos} dataset.}
\vspace{-7pt}
\label{table_proposal_performance}
\begin{center}
\begin{tabular} {c|c}
\hline
\ \ \ \ \ \ \ \textbf{Method} \ \ \ \ \ \ \ & \ \ \ \ \ \ \ \textbf{tIoU=0.8} \ \ \ \ \ \ \ \\
\hline\hline
\ \ \ \ \ \ \ DAP \cite{escorcia2016daps} \ \ \ \ \ \ \ & \ \ \ \ \ \ \ 0.573 \ \ \ \ \ \ \ \\
\hline
\ \ \ \ \ \ \ S-CNN-prop \cite{shou2016temporal} \ \ \ \ \ \ \ & \ \ \ \ \ \ \ 0.524 \ \ \ \ \ \ \ \\
\hline
\ \ \ \ \ \ \ SST \cite{Buch_2017_CVPR} \ \ \ \ \ \ \ & \ \ \ \ \ \ \ 0.672 \ \ \ \ \ \ \ \\
\hline
\ \ \ \ \ \ \ SST (our impl) \ \ \ \ \ \ \ & \ \ \ \ \ \ \ 0.696 \ \ \ \ \ \ \ \\
\hline
\ \ \ \ \ \ \ Backward SST \ \ \ \ \ \ \ & \ \ \ \ \ \ \ 0.684 \ \ \ \ \ \ \ \\ 
\hline
\ \ \ \ \ \ \ Bi-SST \ \ \ \ \ \ \ & \ \ \ \ \ \ \ \textbf{0.711} \ \ \ \ \ \ \ \\
\hline
\end{tabular}
\end{center}
}
\vspace{-20pt}
\end{table}

\subsection{Event Localization}
\noindent \textbf{Metrics.} We use Precision@1000 and Recall@1000 averaged at different tIoU thresholds \{0.3, 0.5, 0.7, 0.9\} as metrics. The evaluation toolkit we used is provided by \cite{Krishna_2017_ICCV}. We also use F1 score to simultaneously consider precision and recall, arguing that F1 is a more reasonable metric for event localization, by showing experimental evidences.

\vspace{5pt}
\noindent \textbf{Compared Methods.} We compare the following methods:
\begin{itemize} [nosep]
\item Random: Both start time and end time are chosen randomly.
\item Forward SST: The method used in~\cite{Buch_2017_CVPR}.
\item Backward SST: Similar as Forward SST, except that the video sequence is fed in a reverse order.
\item Bidirectional SST: Our proposed method. We combine Forward SST and Backward SST and jointly inference by fusing scores for the same predicted proposals.
\end{itemize}

\vspace{5pt}
\noindent \textbf{Settings.} For THUMOS-14 dataset, we follow Buch \textit{et al.} \cite{Buch_2017_CVPR}. For ActivityNet Captions dataset, we do not use multiple strides but with only stride of 64 frames (2 seconds). This gives a reasonable number of unfolding LSTM steps (60 on average) to learn temporal dependency. We do not perform stream sampling and only take the whole video as a single stream, to make sure all ground truth proposals are included. We first train the proposal module (about 5 epochs) to ensure a good initialization and then train the whole model in an end-to-end manner. Adam \cite{kingma2014adam} optimization algorithm with base learning rate of 0.001 and batch size of 1 is used.

\vspace{2pt}
\noindent \textbf{Results.} (1) ActivityNet Captions: As shown in Tab.~\ref{table_event_localization}, Random proposal method gives the highest recall rate among all compared methods. The reason is that most ground truth proposals are pretty long (30\% compared to total video length on average, while only 2\% for THUMOS-14~\cite{jiang2014thumos} action dataset), and thus randomly sampling can possibly cover all ground truth proposals. However, random sampling gives very low precision. A low-precision proposal method will cause performance degeneration for dense captioning system which simply describes all proposals. This is different from action detection, which involves a classification module to further filter out background proposals. Therefore, we mainly refer to F1 score which combines both precision and recall to measure how good the generated proposals are. We compare our bidirectional proposal module with baseline methods using F1 against different tIoU thresholds with ground truth proposals. Our method surpasses SST with clear margins as shown in Tab.~\ref{table_event_localization} and in Fig.~\ref{figure_precision_recall_f1}. This confirms that bidirectional prediction with encoded future context indeed improves proposal quality, compared to single direction prediction model. (2) THUMOS-14: Our proposed Bidirectional SST method can not only be applied to the dense captioning task, but also show superiority on the temporal action proposal task. We set the Non-Maximum Suppression (NMS) threshold to 0.8. We observe in Tab.~\ref{table_proposal_avg_tiou} that our proposed Bi-SST outperforms SST and Backward SST, especially for smaller proposal numbers. In Tab.~\ref{table_proposal_performance}, we compare Bi-SST with other methods. Both Tab.~\ref{table_proposal_avg_tiou} and Tab.~\ref{table_proposal_performance} support that our Bi-SST makes better predictions by combining past and future contexts. It also surpasses other methods and achieves new state-of-the-art results.

\begin{table}
{\footnotesize
\caption{Captioning performance on validation set of ActivityNet Captions using ground truth event proposals.}
\vspace{-3pt}
\label{table_gt_caption_performance}
\begin{center}
\begin{tabular} {c|c}
\hline
\textbf{Method} & \textbf{Meteor}\\
\hline\hline
SST + H & 8.17\\
\hline
Bi-SST + H & 8.68\\
\hline
Bi-SST + E + H & 9.14\\
\hline
Bi-SST + E + H + TDA & 9.36\\
\hline
Bi-SST + E + H + TDA + CG & 9.69\\
\hline
Bi-SST + E + H + TDA + CG + Ranking & \textbf{10.89}\\
\hline
\end{tabular}
\end{center}
}
\vspace{-15pt}
\end{table}

\subsection{Dense Event Captioning}
\label{subsection_dense_captioning}
\noindent \textbf{Metrics.} We mainly refer to Meteor~\cite{banerjee2005meteor} to measure the similarity between two sentences as it is reported to be most correlated to human judgments when a small number of sentence references are given~\cite{vedantam2015cider}. To measure the whole dense captioning system, we average Meteor scores at tIoU thresholds of 0.3, 0.5, 0.7, and 0.9 when describing top 1000 proposals for each video. The same strategy has been adopted in \cite{Krishna_2017_ICCV}. For validation split, we also provide BLEU~\cite{papineni2002bleu}, Rouge-L~\cite{lin2004rouge}, and CIDEr-D~\cite{vedantam2015cider} scores for complete comparison. For test split, we report Meteor score, since the test server only returns Meteor result.

\begin{table*}[!htb]
{\footnotesize
\caption{Performance of different methods on ActivityNet Captions validation set. All values are reported in percentage (\%). No validation result is provided by \cite{Krishna_2017_ICCV}. H: context vectors, E: event features, TDA: temporal dynamic attention, CG: context gate.}
\vspace{-7pt}
\label{table_dense_captioning_val}
\begin{center}
\begin{tabular} {c|c|c|c|c|c|c|c}
\hline
\textbf{Method} & \textbf{BLEU-1} & \textbf{BLEU-2} & \textbf{BLEU-3} & \textbf{BLEU-4} & \textbf{Meteor} & \textbf{Rouge-L} & \textbf{CIDEr-D}\\
\hline\hline
SST + H & 16.78 & 5.94 & 2.22 & 0.88 & 7.87 & 16.75 & 8.17\\
\hline\hline
Bi-SST + H & 17.25 & 6.48 & 2.68 & 1.20 & 8.35 & 17.56 & 8.49\\
Bi-SST + E & 17.51 & 7.17 & 3.08 & 1.32 & 8.36 & 17.96 & 9.13 \\
Bi-SST + E + H & 17.50 & 6.95 & 2.94 & 1.28 & 8.78 & 17.68 & 9.10 \\
Bi-SST + E + H + TDA & 18.70 & 8.17 & 3.63 & 1.59 & 9.00 & 18.64 & 10.02\\
Bi-SST + E + H + TDA + CG & \textbf{19.37} & 8.69 & 4.03 & 1.89 & 9.19 & \textbf{19.29} & 11.03 \\ 
Bi-SST + E + H + TDA + CG + Ranking & 18.99 & \textbf{8.84} & \textbf{4.41} & \textbf{2.30} & \textbf{9.60} & 19.10 & \textbf{12.68}\\
\hline
\end{tabular}
\end{center}
}
\vspace{-15pt}
\end{table*}

\begin{table}
{\footnotesize
\caption{Comparison with the state-of-art method on ActivityNet Captions test set. The test server returns only Meteor score (in percentage (\%)).}
\vspace{-7pt}
\label{table_dense_captioning_server}
\begin{center}
\begin{tabular} {c|c}
\hline
\textbf{Method} & \textbf{Meteor} \\
\hline\hline
\ \ \ \ \ \ \ \ \ Krishna \emph{et al.}~\cite{Krishna_2017_ICCV} \ \ \ \ \ \ \ \ \ & \ \ \ \ \ \ \ \ \ 4.82 \ \ \ \ \ \ \ \ \ \\
\hline
Ours & 9.65 \\
\hline
\end{tabular}
\end{center}
}
\vspace{-20pt}
\end{table}

\vspace{5pt}
\noindent \textbf{Compared Methods.} We denote ``H'' as context vectors, ``E'' as event clip features, ``TDA'' as temporal dynamic attention fusion, and ``CG'' as context gate, respectively. We compare the following methods:
\begin{itemize} [nosep]
\item SST + H: This method utilizes SST \cite{Buch_2017_CVPR} to generate proposals and represents them with corresponding hidden states for generating descriptions. This approach is served as our baseline.
\item Bi-SST + H: We apply our bidirectional proposal method to generating proposals. The hidden states from both direction are concatenated to represent an event.
\item Bi-SST + E: Mean pooled event feature is used to represent the detected event.
\item Bi-SST + E + H: Mean pooled event feature and hidden states are concatenated for representation.
\item Bi-SST + E + H + TDA: Temporal dynamic attention (TDA) is used to dynamically construct visual input to the decoder.
\item Bi-SST + E + H + TDA + CG: Context gate is used to balance the attended event feature and contexts.
\item Bi-SST + E + H + TDA + CG + Ranking: Joint ranking is further applied in inference time.
\end{itemize}


\vspace{5pt}
\noindent \textbf{Results.} The results of our methods and the baseline approach on the ActivityNet Captions validation split are provided in Tab.~\ref{table_gt_caption_performance} and in Tab.~\ref{table_dense_captioning_val}. We can see that, our 6 variants all outperform the baseline method with large margins. 

Compared to the baseline SST + H, our bidirectional model (Bi-SST + H) gives better performance when captioning 1000 proposals. This verifies that considering both past and future event context also help improve describing an event. 

Combining both event clip features and context vectors (Bi-SST+E+H) is better than event clip features (Bi-SST+E) or context vectors (Bi-SST+H) alone. We mainly refer to Meteor score for comparison, as it shows better consistency with human judgments with a small number of reference sentences (in our case, only one reference sentence). We notice there is slight inconsistency for other metrics, which has also been observed by \cite{yao2015describing,Krishna_2017_ICCV}. This is caused by the imperfection of sentence similarity measurement.

Based on the results of Bi-SST+E+H+TDA, applying attention mechanism instead of mean pooling to dynamically fuse event clip features and context vectors further improves all scores. This variant performs better as it can generates more semantic related word by attending on video features at each decoding step.
Combining context gating function further boosts the performance with clear margins. This supports that explicitly modeling the relative contribution from event features and contexts in decoding time help better describe the event.
Using joint ranking at inference time further improves the whole system, as it gives more confident predictions on both event proposals and corresponding descriptions.

In Tab.~\ref{table_dense_captioning_server}, comparison of our system with the state-of-the-art method \cite{Krishna_2017_ICCV} is presented. Note that our approach uses only C3D features and does not involve any extra data. While totally comparable to Krishna \emph{et al.} \cite{Krishna_2017_ICCV}, our method surpasses \cite{Krishna_2017_ICCV} with \textbf{100\%} performance gain. This strongly supports the effectiveness of our proposed model.

\begin{figure}[t]
\begin{center}
\includegraphics [width=1.0\linewidth]{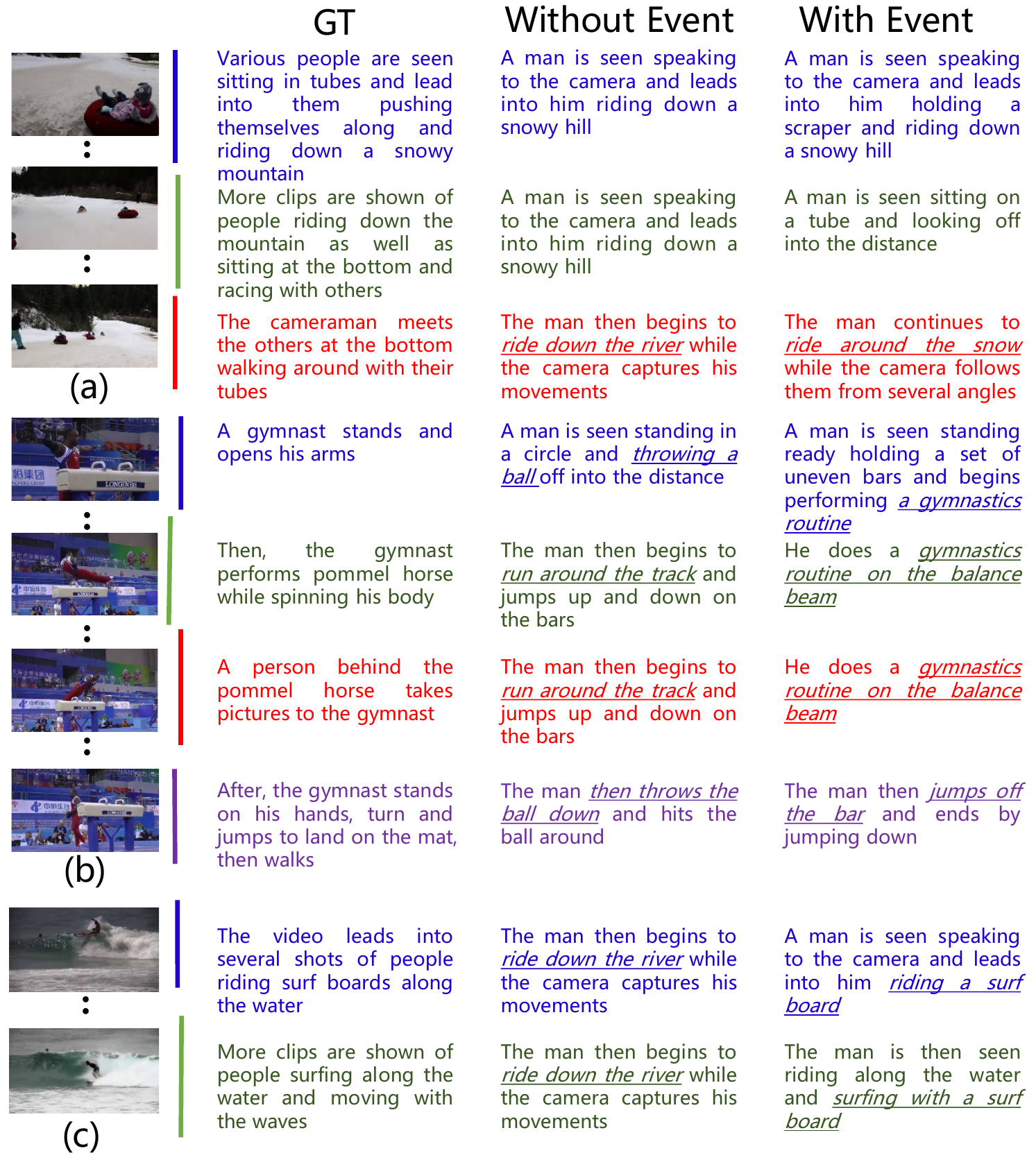}
\caption{Qualitative dense-captioning analysis for model without or with event clip fusion. The underline words are important details for comparison. Note that we only show proposals with maximum tIoU with the ground truths. (Best viewed in color)}
\label{figure_case}
\end{center}
\vspace{-10pt}
\end{figure}

\vspace{5pt}
\noindent \textbf{Qualitative Analysis.} For intuitively analyzing the effectiveness of fusing event clip for dense captioning, we show some cases in Fig. \ref{figure_case}. The fusion mechanism allows the system to pay more ``attention" to current event while simultaneously referring to contexts, and thus can generate more semantic-related sentences. In contrast, the system without event clip fusion generally tends to make more semantic mistakes, either incorrect (Fig. \ref{figure_case} (a) and (b)) or semantic ambiguous (Fig. \ref{figure_case} (c)). For example, when describing video (c), by incorporating event clip features, the system is more confident to say ``The man is surfing with a surf board'' instead of simply saying ``riding down the river.'' 

\begin{figure}
\begin{center}
\includegraphics[width=1.0\linewidth]{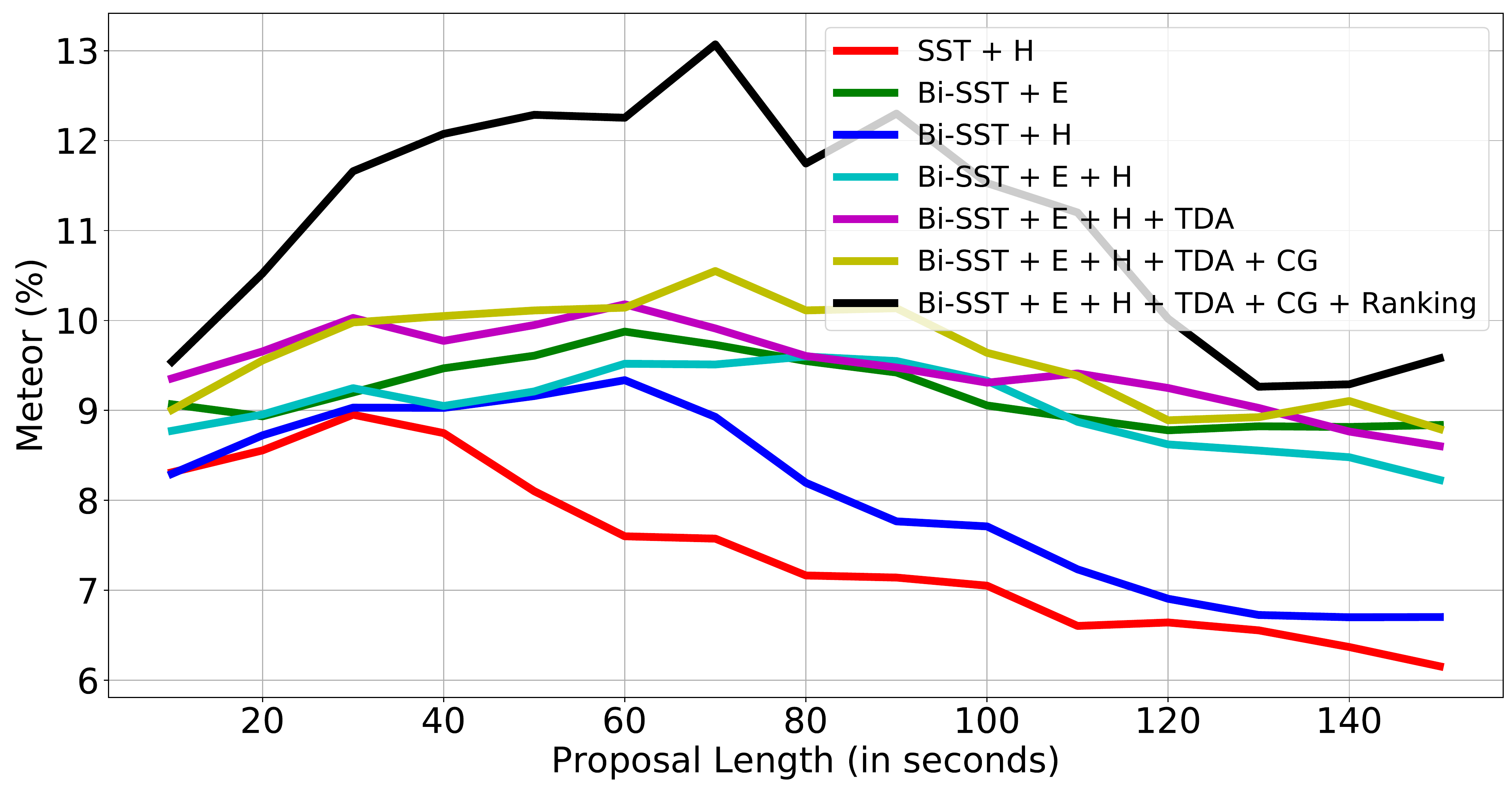}
\caption{Meteor scores \emph{vs} event proposal lengths.}
\label{figure_meteor_proposal}
\end{center}
\vspace{-10pt}
\end{figure}

Fig. \ref{figure_meteor_proposal} shows how Meteor scores change as proposal lengths vary from a few seconds to several minutes. We can see that the performances of all methods degenerate when describing very long proposals ($>$ 60s). This suggests that understanding long events and describing them is still a challenging problem, as long events are usually more complicated. Bi-SST+H works better than SST-H as we combine both past and future context information. We note that SST+H and Bi-SST+H both go down steeply as proposals become longer. The reason is that it is still very hard for LSTM to learn long-term dependency. Using only hidden states to represent an event is thus quite suboptimal. In contrast, fusing event features compensates such information loss. All methods using ``E'' (event features) show much better performance than their counterparts. Besides, our model with joint ranking further improves the performance of the whole system with large margins.

\section{Conclusion}
In this paper we identified and handled two challenges on the task of dense video captioning, which are context fusion and event representation. We proposed a novel bidirectional proposal framework, namely, Bidirectional SST, to encode both past and future contexts, with the motivation that both past and future contexts help better localize the current event. Building on this proposal module, we further reused the proposal hidden states as context vectors and dynamically fused with event clip features to generate the visual representation. The proposed system can be trained in an end-to-end manner. The extensive quantitative and qualitative experimental results demonstrate the superiority of our model in both localizing events and describing them.

\section{Supplementary Material}

\subsection{More Qualitative Results for Dense Captioning}
\begin{figure*}
\centering
\includegraphics[width=1.0 \linewidth]{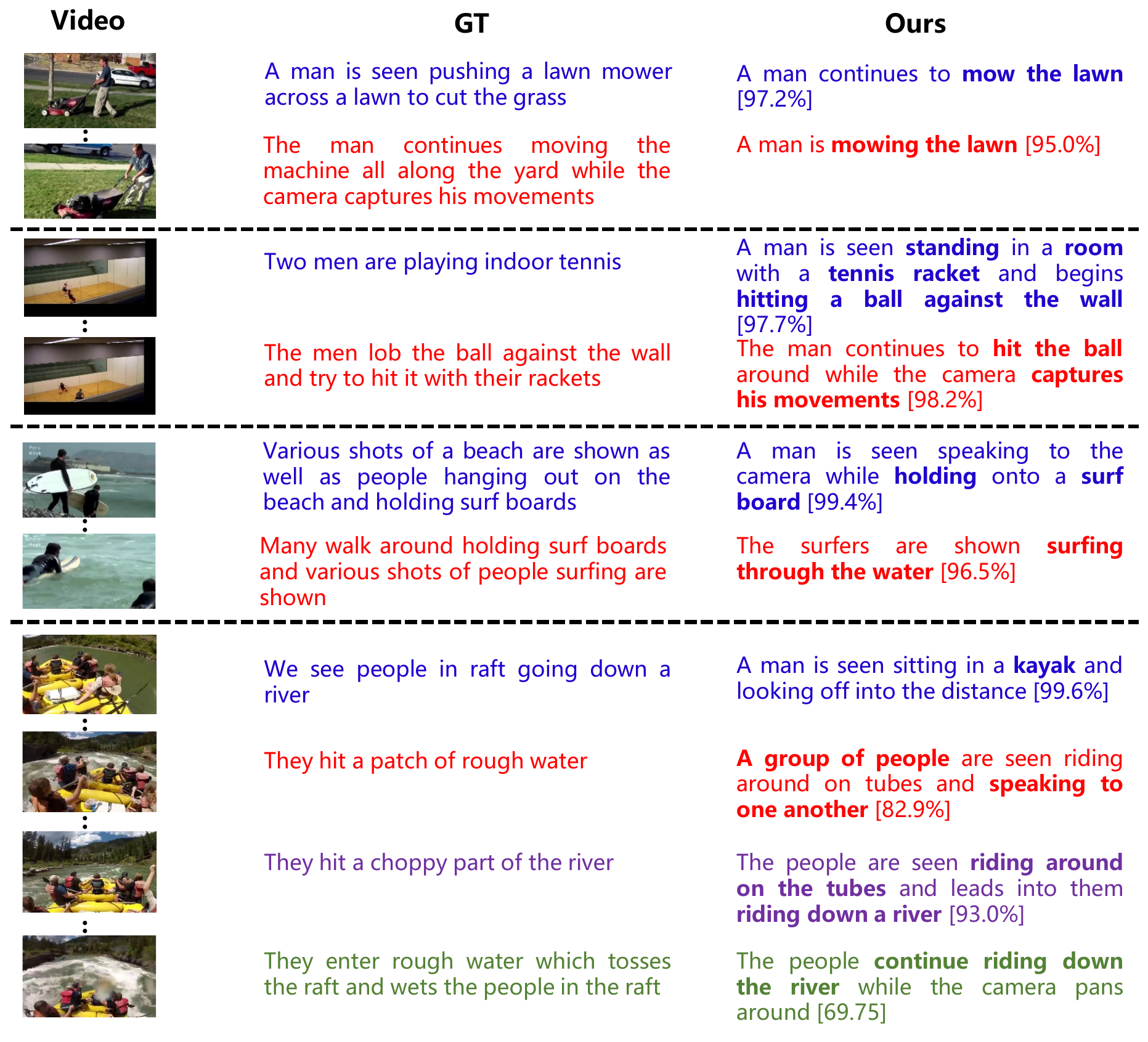}
\caption{Dense captioning qualitative examples. Numbers in the brackets are tIoUs between the predicted proposals and the corresponding ground truths. Note that we show proposals with max tIoU with the ground truths.}
\label{figure_case_2}
\end{figure*}

\begin{figure*}
\centering
\includegraphics[width=1.0 \linewidth]{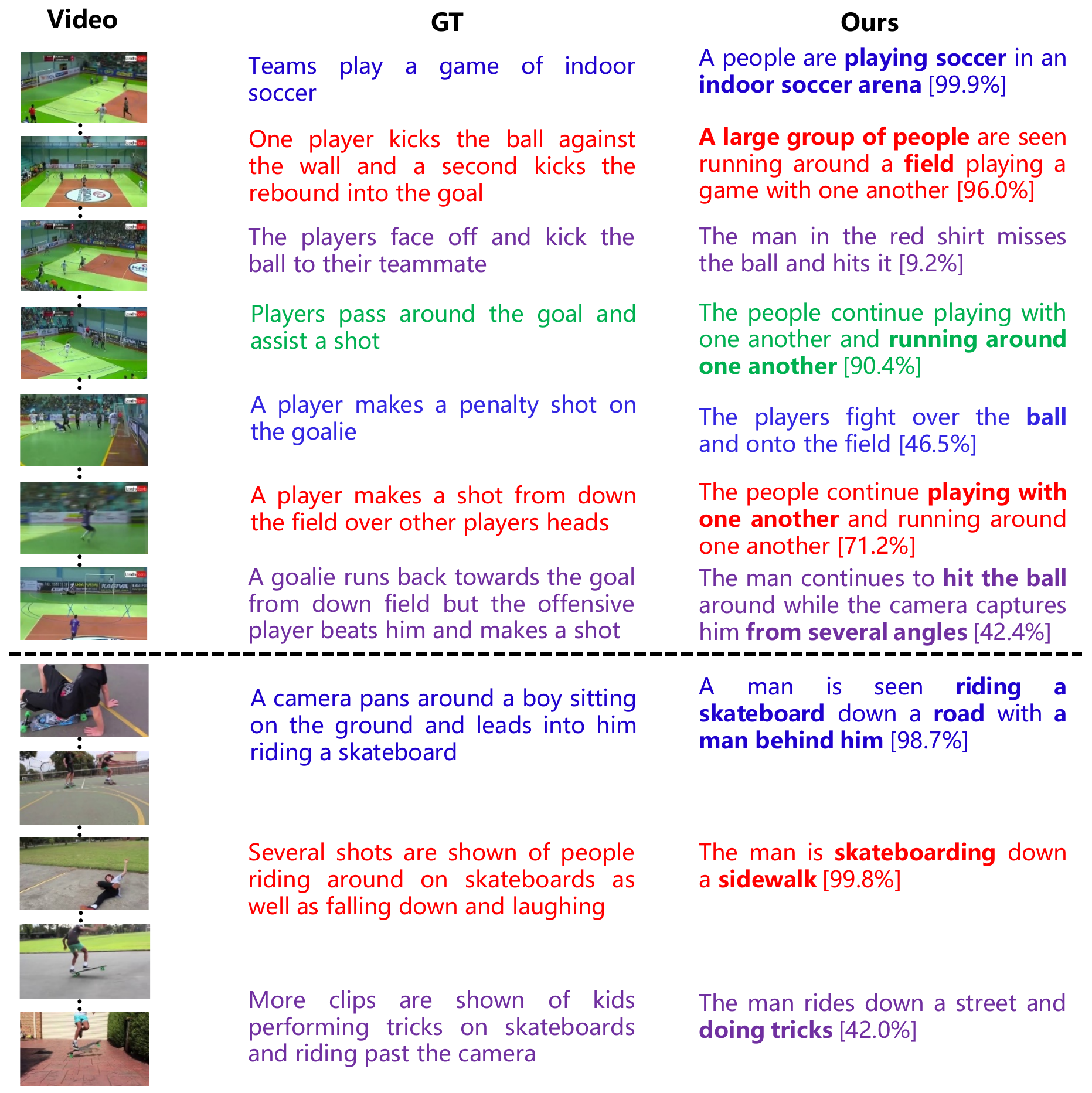}
\caption{Dense captioning qualitative examples. Numbers in the brackets are tIoUs between the predicted proposals and the corresponding ground truths. Note that we show proposals with max tIoU with the ground truths.}
\label{figure_case_1}
\end{figure*}

In Fig.~\ref{figure_case_2} and Fig.~\ref{figure_case_1}, we provide more qualitative results for our best dense captioning model ``Bi-SST+E+H+TDA+CG+Ranking.'' Note that our generated captions even have more details than the ground truths in many cases.

\subsection{Captioning Performance on Different Activity Categories}
\begin{figure*}
\centering
\includegraphics[width=0.95 \linewidth]{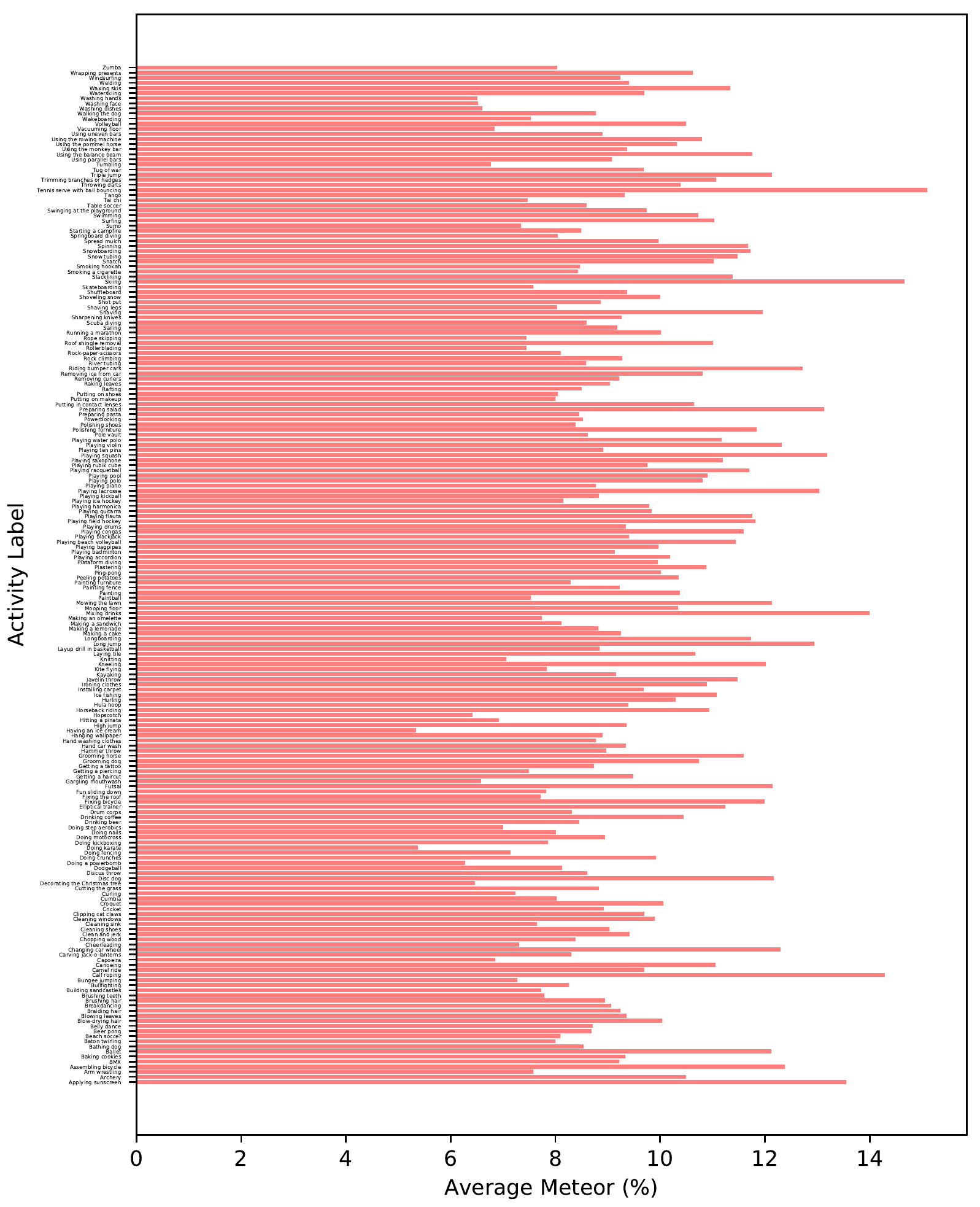}
\caption{Dense-captioning performance on different activity labels for our best model ``Bi-SST+E+H+TDA+CG+Ranking''.}
\label{figure_meteor_action_label}
\end{figure*}

In Fig. \ref{figure_meteor_action_label}, we provide detailed dense captioning performance for videos from different activity categories. The top 5 best-performing categories are ``Tennis serve with ball bouncing'' (Meteor: 15.1), ``Skiing'' (Meteor: 14.7), ``Calf roping'' (Meteor: 14.3), ``Mixing drinks'' (Meteor: 14.0), ``Applying sunscreen'' (Meteor: 13.6). The top 5 worst-performing categories are ``Having an ice cream'' (Meteor: 5.3), ``Doing Karate'' (Meteor: 5.4), ``Doing a powerbomb'' (Meteor: 6.3), ``Hopscotch'' (Meteor: 6.4), ``Decorating the Christmas tree'' (Meteor: 6.5). The different performances among different activity categories could possibly be attributed to varied video durations, varied complexity of videos, varied annotation qualities, and so on.

\vspace{10pt}
\noindent \textbf{Acknowledgement} The author Yong Xu would like to thank the supports by National Nature Science Foundation of China (U1611461 and 61672241), the Cultivation Project of Major Basic Research of NSF-Guangdong Province (2016A030308013).

\newpage

{\footnotesize
\bibliographystyle{ieee}
\bibliography{egbib}
}

\end{document}